\documentclass[twoside]{article}

\usepackage[accepted]{general}
\usepackage{sec/package}
\input{sec/def.set}
\usepackage{sec/bibstyle}
\begin{document}

%

%
\runningauthor{Guo, Jin, Wang, Qiu, Zhang, Zhu, Zhang, Wipf}

\twocolumn[

\aistatstitle{Fork or Fail: Cycle-Consistent Training with Many-to-One Mappings}

\aistatsauthor{ Qipeng Guo\textsuperscript{1,2} \And Zhijing Jin\textsuperscript{1,3} \And Ziyu Wang\textsuperscript{4} 
\And Xipeng Qiu\textsuperscript{2} }
\aistatsauthor{ Weinan Zhang\textsuperscript{5} \And Jun Zhu\textsuperscript{4} \And  Zheng Zhang\textsuperscript{1} \And David Wipf\textsuperscript{1}}
\begin{center}
\textsuperscript{1}Amazon Shanghai AI Lab { } 
\textsuperscript{2}Fudan University { } 
\textsuperscript{3}Max Planck Institute for Intelligent Systems, Tübingen
\\
\textsuperscript{4}Tsinghua University { }
\textsuperscript{5}Shanghai Jiao Tong University
\end{center}
\aistatsaddress{}
]

\begin{abstract}
\vspace*{-0.1cm}
Cycle-consistent training is widely used for jointly learning a forward and inverse mapping between two domains of interest without the cumbersome requirement of collecting matched pairs within each domain.  In this regard, the implicit assumption is that there exists (at least approximately) a ground-truth bijection such that a given input from either domain can be accurately reconstructed from successive application of the respective mappings.  But in many applications no such bijection can be expected to exist and large reconstruction errors can compromise the success of cycle-consistent training.  As one important instance of this limitation, we consider practically-relevant situations where there exists a many-to-one or surjective mapping between domains.  To address this regime, we develop a conditional variational autoencoder (CVAE) approach that can be viewed as converting surjective mappings to implicit bijections whereby reconstruction errors in both directions can be minimized, and as a natural byproduct, realistic output diversity can be obtained in the one-to-many direction.  As theoretical motivation, we analyze a simplified scenario whereby minima of the proposed CVAE-based energy function align with the recovery of ground-truth surjective mappings.  On the empirical side, we consider a synthetic image dataset with known ground-truth, as well as a real-world application involving natural language generation from knowledge graphs and vice versa, a prototypical surjective case.  For the latter, our CVAE pipeline can capture such many-to-one mappings during cycle training while promoting textural diversity for graph-to-text tasks.\footnote{Our code is available \codeURL{}.}



\end{abstract}

\section{Introduction}
Given data $\bx \in \calX$ from domain $\calX$ and $\by \in \calY$ from domain $\calY$, it is often desirable to learn bidirectional mappings $f: \calY \rightarrow \calX$ and $g: \calX \rightarrow \calY$ such that for matched pairs $\{\bx,\by\}$, we have that $\bx \approx \hat{\bx} \triangleq f(\by)$ and $\by \approx \hat{\by} \triangleq g(\bx)$.  When provided with a corpus of suitably aligned data, this amounts to a straightforward supervised learning problem.  However, in many applications spanning computer vision \citep{DBLP:conf/iccv/ZhuPIE17}, natural language processing \citep{DBLP:conf/iclr/LampleCDR18,DBLP:conf/iclr/ArtetxeLAC18} and speech recognition \citep{DBLP:conf/icassp/HoriAHZWR19}, we may only have access to individual samples from $\calX$ and $\calY$ but limited or no labeled ground-truth matches between domains, since, for example, the labeling process may be prohibitively expensive. To address this commonly-encountered situation, cycle-consistent training represents an unsupervised means of jointly learning $f$ and $g$ by penalizing the cycle-consistency reconstruction losses $\left\|\bx - f[ g(\bx) ] \right\|$ and $\left\|\by - g[ f(\by) ] \right\|$ using non-parallel samples from $\calX$ and $\calY$ and some norm or distance metric $\| \cdot \|$ \citep{DBLP:conf/iccv/ZhuPIE17}.

However, this process implicitly assumes that there exists a suitable bijection between domains (implying $f = g^{-1}$ and $g = f^{-1}$), an assumption that frequently does not hold for practical applications of cycle-consistent training.  As a representative example related to natural language understanding, each $\bx$ may represent a text segment while $\by$ corresponds with the underlying knowledge graph describing the text content.  The relationship between these domains is surjective, but not bijective, in the sense that multiple sentences with equivalent meaning but different syntactic structure can be mapped to the same knowledge graph.  Hence if we follow any possible learned mapping $\bx \rightarrow \hat{\by} \rightarrow \hat{\bx}$, there will often be significant error between $\bx$ and the reconstructed $\hat{\bx}$.  In other words, no invertible transformation exists between domains and there will necessarily be information about $\bx$ that is lost when we map through $\calY$ space.  Additionally, deterministic mappings do not reflect the ground-truth conditional distribution $p_{gt}(\bx|\by)$, which is necessary for the generation of diverse text consistent with a given knowledge graph.

Despite these limitations, there has been relatively little effort or rigorous analysis devoted to explicitly addressing the lack of a bijection in applications of cycle-consistent training; Section \ref{sec:related_work} on related work will discuss this point in greater detail.  As a step towards filling this void, in Section \ref{sec:model_development} we will consider replacing the typical deterministic cycle training pipeline with a stochastic model reflecting $p_{gt}(\bx|\by)$ and $p_{gt}(\by|\bx)$ for the $\bx \rightarrow \hat{\by} \rightarrow \hat{\bx}$ and $\by \rightarrow \hat{\bx} \rightarrow \hat{\by}$ cycles respectively.  In doing so, we apply a conditional variational autoencoder (CVAE) formulation \citep{doersch2016tutorial,sohn2015learning} to deal with the intractable integrals that arise.  Note that although the proposed CVAE methodology can be generalized, we will herein restrict ourselves to situations where there exists a many-to-one mapping from $\bx$ to $\by$ (i.e., a surjection) as originally motivated by our interest in conversions between knowledge graphs and diverse, natural language text.

Proceeding further, Section \ref{sec:analysis} provides theoretical support by analyzing a simplified scenario whereby minima of the proposed CVAE-based energy function align with the recovery of ground-truth surjective mappings.  To the best of our knowledge, this is
 the only demonstration of a cycle-consistent model with any type of performance guarantee within a non-trivial, non-bijective context.  We then turn to empirical validation in Section \ref{sec:experiments} that corroborates our theory via a synthetic image example and demonstrates real-world practicality on an application involving the conversion between diverse natural language and knowledge graphs taken from the WebNLG dataset.   Overall, experimental results indicate that our proposed CVAE pipeline can approximate surjective mappings during cycle training, with performance on par with supervised alternatives, while promoting diversity for the graph-to-text direction.

\vspace*{-0.2cm}
\section{Related Work
\vspace*{-0.2cm}
} \label{sec:related_work}

\paragraph{General Cycle-Consistent Training}
The concept of leveraging the transitivity of two functions that serve as inverses to one another has been applied to a variety of tasks.  For example, in computer vision,  forward-backward consistency has been used extensively in computer vision \citep{DBLP:conf/icpr/KalalMM10,DBLP:conf/eccv/SundaramBK10}, and cycle-consistent training pipelines underlie image style transfer \citep{DBLP:conf/iccv/ZhuPIE17,liu2017unsupervised}, depth estimation \citep{DBLP:conf/cvpr/GodardAB17}, and unsupervised domain adaptation \citep{hoffman2018cycada} pipelines. Turning to natural language processing (NLP), back translation \citep{DBLP:conf/acl/SennrichHB16,DBLP:conf/emnlp/EdunovOAG18,jin2020simple} and dual learning \citep{DBLP:conf/acl/ChengXHHWSL16,DBLP:conf/nips/HeXQWYLM16} have been widely deployed for unsupervised machine translation. Similar techniques have also contributed to applications such as language style transfer \citep{DBLP:conf/nips/ShenLBJ17,DBLP:conf/emnlp/JinJMMS19}.  However, the above models primarily rely on deterministic pipelines that implicitly assume a bijection even if one does not actually exist. 

And finally, a VAE-inspired model for converting between knowledge graphs and text is considered in \citep{tseng2020generative}.  But again there is no explicit accounting for non-bijective data as a shared latent space is assumed to contain all information from \emph{both} $\bx$ and $\by$ domains, and the proposed model is designed and tested for semi-supervised learning (not fully unsupervised cycle-consistency).  Moreover, for tractable inference, some terms from the variational bound on the log-likelihood (central to all VAE models) are heuristically removed; hence the relationship with the original, motivating probabilistic model remains unclear.


\paragraph{Non-Bijective Mappings}

Non-bijective mappings are investigated in applications such as
multi-domain image-to-image translation \citep{choi2018stargan}, voice conversion \citep{kameoka2018stargan}, multi-attribute text style transfer \citep{lample2018multiple}, music transfer \citep{bitton2018modulated}, and multi-modal generation \citep{shi2019variational}. Most of this work uses adversarial neural networks, or separate decoders \citep{lee2019many,mor2018universal}, and  one case even applies a CVAE model \citep{jha2018disentangling}. However, all the above assume multiple \emph{pre-defined} style domains and require data be clearly separated a priori according to these domains to train non-bijective mappings. In contrast, our proposed  model assumes a completely arbitrary surjective mapping that can be learned from the data without such additional domain-specific side information pertaining to styles or related (so ours can fit unknown styles mixed within an arbitrary dataset).  One exception is \citep{zhu2017toward}, which handles general non-bijective image mappings using a hybrid VAE-GAN model but unlike our approach, it requires matched $\{\bx,\by\}$ pairs for training. 





\vspace*{-0.2cm}
\section{Model Development}\label{sec:model_development}
\vspace*{-0.2cm}

We will first present a stochastic alternative to deterministic cycle-consistency that, while useful in principle for handling surjective (but explicitly non-bijective) mappings, affords us with no practically-realizable inference procedure.  To mitigate this shortcoming, we then derive a tractable CVAE approximation and discuss some of its advantages.  Later in Section \ref{sec:analysis} we will analyze the local and global minima of this cycle-consistent CVAE model in the special case where the decoder functions are restricted to being affine.

\subsection{Stochastic Cycle-Consistent Formulation} \label{sec:stochastic_formulation}

Although the proposed methodology can be generalized, we will herein restrict ourselves to situations where there exists a many-to-one mapping from $\bx$ to $\by$ (i.e., a surjection) and the resulting asymmetry necessitates that the $\bx \rightarrow \hat{\by} \rightarrow \hat{\bx}$ and $\by \rightarrow \hat{\bx} \rightarrow \hat{\by}$ cycles be handled differently.  In this regard, our starting point is to postulate an additional latent variable $\bu \in \calU$ that contributes to a surjective matched pair $\{\bx,\by\}$ via
\begin{equation} \label{eq:gt_surjection}
\bx = h_{gt}\left(\by,\bu \right) ~~~ \mbox{and} ~~~ \by = h^+_{gt} \left(\bx \right),
\end{equation}
where $h_{gt}: \calY \times \calU \rightarrow \calX$ and $h^+_{gt} : \calX \rightarrow \calY$ represent ground-truth mappings we would ultimately like to estimate.  For this purpose we adopt the approximations $h_\theta : \calY \times \calZ \rightarrow \calX$ and $h_\theta^+ : \calX \rightarrow \calY$ with trainable parameters $\theta$, noting that the second input argument of $h_\theta$ is now $\bz \in \calZ$ instead of $\bu \in \calU$.  This is because the latter is unobservable and it is sufficient to learn a mapping that preserves the surjection between $\bx$ and $\by$ without necessarily reproducing the exact same functional form of $h_{gt}$.  For example, if hypothetically $\bu = \pi(\bz)$ in (\ref{eq:gt_surjection}) for some function $\pi$, we could redefine $h_{gt}$ as a function of $\bz$ without actually changing the relationship between $\bx$ and $\by$.

We are now prepared to define a negative conditional log-likelihood loss for both cycle directions, averaged over the distributions of $\by$ and $\bx$ respectively.  For the simpler $\by \rightarrow \hat{\bx} \rightarrow \hat{\by}$ cycle, we define
\begin{equation} \label{eq:y_likelihood}
\ell_y(\theta) ~ \triangleq ~ - \int \left(\log \int p_\theta\left(\by | \hat{\bx}  \right) p(\bz)d \bz \right) \rho_{gt}^y(d\by),
\end{equation}
where $\hat{\bx} = h_\theta \left( \by, \bz \right)$, $p_\theta\left(\by | \hat{\bx}  \right)$ is determined by $h_\theta^+$ and an appropriate domain-specific distribution, and $p(\bz)$ is assumed to be fixed and known (e.g., a standardized Gaussian).  Additionally, $\rho_{gt}^y$ denotes the ground-truth probability measure associated with $\calY$.  Consequently, $\rho_{gt}^y(d\by)$ is the measure assigned to the infinitesimal $d\by$, from which it obviously follows that $\int \rho_{gt}^y(d\by) = 1$.  Note that the resulting derivations can apply even if no ground-truth density $p_{gt}(\by)$ exists, e.g., a counting measure over training samples or discrete domains can be assumed within this representation.

Given that there should be no uncertainty in $\by$ when conditioned on $\bx$, we would ideally like to learn parameters whereby $p_\theta\left(\by | \hat{\bx}  \right)$, and therefore $\ell_y(\theta)$, degenerates while reflecting a many-to-one mapping.  By this we mean that
\begin{equation}
\by \approx \hat{\by} = h^+_\theta(\hat{\bx}) = h^+_\theta\left(h_\theta\left[\by,\bz \right]\right), ~~~ \forall \bz \sim p(\bz).
\end{equation}
Hence the $\by \rightarrow \hat{\bx} \rightarrow \hat{\by}$ cycle only serves to favor (near) perfect reconstructions of $\by$ while ignoring any $\bz$ that can be drawn from $p(\bz)$.  The latter stipulation is unique relative to typical deterministic cycle-consistent training, which need not learn to ignore randomness from an additional confounding latent factor.

In contrast, the $\bx \rightarrow \hat{\by} \rightarrow \hat{\bx}$ cycle operates somewhat differently, with the latent $\bz$ serving a more important, non-degenerate role.  Similar to before, we would ideally like to minimize the negative conditional log-likelihood given by
\begin{equation} \label{eq:x_likelihood}
\ell_x(\theta) ~ \triangleq ~ -  \int \left( \log\int p_\theta\left(\bx | \hat{\by}, \bz \right) p(\bz)d \bz \right) \rho_{gt}^x(d\bx),
\end{equation}
where now $\hat{\by} = h_\theta^+\left( \bx \right)$, $p_\theta\left(\bx | \by, \bz \right)$ depends on $h_\theta$, and $\rho_{gt}^x$ represents the ground-truth probability measure on $\calX$.  Here $\hat{\by}$ can be viewed as an estimate of all the information pertaining to the unknown paired $\by$ as preserved through the mapping $h_\theta^+$ from $\bx$ to $\by$ space.  Moreover, if cycle-consistent training is ultimately successful, both $\hat{\by}$ and $\by$ should be independent of $\bz$, and it should be the case that $\int p_\theta\left(\bx | \hat{\by},\bz \right) p(\bz)d \bz = p_\theta\left(\bx | \hat{\by} \right) \approx p_{gt}\left(\bx | \by \right)$.

Per these definitions, it is also immediately apparent that $p_\theta\left(\bx | \hat{\by} \right)$ is describing the distribution of $h_\theta \left(\by,\bz\right)$ conditioned on $\by$ being fixed to $h_\theta^+\left( \bx \right)$.  This can be viewed as a stochastic version of the typical $\bx \rightarrow \hat{\by} \rightarrow \hat{\bx}$ cycle, whereas now the latent $\bz$ allows us to spread probability mass across all $\bx$ that are consistent with a given $\hat{\by}$.  In fact, if we set $\bz = \bz'$ to a fixed null value (i.e., change $p(\bz)$ to a Dirac delta function centered at some arbitrary $\bz'$), we recover this traditional pipeline exactly, with $-\log p_\theta(\bx| \by, \bz')$ simply defining the implicit loss function, with limited ability to assign high probability to multiple different values of $\bx$ for any given $\by$.



\subsection{CVAE Approximation} \label{sec:cvae_model}

While $\ell_y(\theta)$ from Section \ref{sec:stochastic_formulation} can be efficiently minimized using stochastic sampling from $p(\bz)$ to estimate the required integral, the $\ell_x(\theta)$ term is generally intractable.  Note that unlike $\ell_y(\theta)$, directly sampling $\bz$ is not a viable solution for $\ell_x(\theta)$ since $p_\theta(\bx|\hat{\by},\bz)$ can be close to zero for nearly all values of $\bz$, and therefore, a prohibitively large number of samples would be needed to obtain reasonable estimates of the integral.  Fortunately though, we can form a trainable upper bound using a CVAE architecture that dramatically improves sample efficiency \citep{doersch2016tutorial,sohn2015learning}.  Specifically, we define
\begin{eqnarray} \label{eq:CVAE_objective}
\ell_x(\theta,\phi) & \triangleq & \int \Big\{ - \mathbb{E}_{q_{\tiny \phi}\left(\bz |\bx \right)} \left[\log p_{\tiny \theta} \left(\bx | \hat{\by}, \bz  \right)  \right]p(\bz)d \bz  \nonumber \\
&&  ~+ ~~  \mathbb{KL}\left[q_\phi(\bz|\bx)\|p(\bz)\right] \Big\} \rho_{gt}^x(d\bx),
\end{eqnarray}
where in this context, $p_{\tiny \theta} \left(\bx | \hat{\by}, \bz  \right)$ is referred to as a \emph{decoder} distribution while $q_{\tiny \phi}\left(\bz |\bx \right)$ represents a trainable \emph{encoder} distribution parameterized by $\phi$.  And by design of general VAE-based models, we have that $\ell_x(\theta,\phi) \geq  \ell_x(\theta)$ for all $\phi$ \citep{kingma2014vae,Rezende2014}.  Note that we could also choose to condition $q_{\tiny \phi}\left(\bz |\bx \right)$ and $p(\bz)$ on $\hat{\by}$, although this is not required to form a valid or maximally-tight bound.  In the case of $q_{\tiny \phi}\left(\bz |\bx \right)$, $\hat{\by}$ is merely a function of $\bx$ and therefore contains no additional information beyond direct conditioning on $\bx$ (and the stated bound holds regardless of what we choose for $q_\phi$).  In contrast, $p(\bz)$ defines the assumed generative model which we are also free to choose; however, conditioning on $\hat{\by}$ can be absorbed into $p_{\tiny \theta} \left(\bx | \hat{\by}, \bz \right)$ such that there is no change in the representational capacity of $p_{\tiny \theta} \left(\bx | \hat{\by} \right) = \int p_{\tiny \theta} \left(\bx | \hat{\by}, \bz  \right) p(\bz) d \bz$.

Given (\ref{eq:CVAE_objective}) as a surrogate for $\ell_x(\theta)$ and suitable distributional assumptions, the combined cycle-consistent loss
\begin{equation} \label{eq:basic_cycle_loss}
\ell_{\tiny \mathrm{cycle}}(\theta,\phi) ~ = ~ \ell_x(\theta,\phi) + \ell_y(\theta)
\end{equation}
can be minimized over $\{\theta,\phi\}$ using stochastic gradient descent and the reparameterization trick from \citep{kingma2014vae,Rezende2014}.  We henceforth refer to this formulation as \textit{CycleCVAE}.  And as will be discussed in more detail later, additional constraints, regularization factors, or inductive biases can also be included to help ensure identifiability of ground-truth mappings.  For example, we may consider penalizing or constraining the divergence between the distributions of $\by$ and $\hat{\by} = h_\theta^+(\bx)$, both of which can be estimated from unpaired samples from $\rho_{gt}^y$ and $\rho_{gt}^x$ respectively.  This is useful for disambiguating the contributions of $\by$ and $\bz$ to $\bx$ and will be equal to zero (or nearly so) if $h_\theta^+ \approx h_{gt}^+$ (more on this in Section \ref{sec:analysis} below).

\subsection{CycleCVAE Inference}

Once trained and we have obtained some optimal CycleVAE parameters $\{\theta^*,\phi^*\} \approx \arg\min_{\theta,\phi} \ell_{\tiny \mathrm{cycle}}(\theta,\phi)$, we can compute matches for test data in either the $\bx_{test} \rightarrow \by_{test}$ or $\by_{test} \rightarrow \bx_{test}$ direction.  For the former, we need only compute $\hat{\by}_{test} = h^+_{\theta^*} (\bx_{test})$ and there is no randomness involved.  In contrast, for the other direction (one-to-many) we can effectively draw approximate samples from the posterior distribution $p_{\theta^*}(\bx|\by_{test})$ by first drawing samples $\bz \sim p(\bz)$ and then computing $\hat{\bx}_{test} = h_{\theta^*}(\by_{test},\bz)$.

\subsection{CycleCVAE Advantages in Converting Surjections to Implicit Bijections} \label{sec:cvae_advantages}

Before proceeding to a detailed theoretical analysis of CycleCVAE, it is worth examining a critical yet subtle distinction between CycleCVAE and analogous, deterministic baselines.  In particular, given that the  $\bx \rightarrow \hat{\by} \rightarrow \hat{\bx}$ cycle will normally introduce reconstruction errors because of the lack of a bijection as discussed previously, we could simply augment traditional deterministic pipelines with a $\bz$ such that $\bx \rightarrow \{\hat{\by},\bz \} \rightarrow \hat{\bx}$ forms a bijection.  But there remain (at least) two unresolved problems.  First, it is unclear how to choose the dimensionality and distribution of $\bz \in \calZ$ such that we can actually obtain a bijection.  For example, if $\mbox{dim}[\calZ]$ is less than the unknown $\mbox{dim}[\calU]$, then the reconstruction error $\| \bx - \hat{\bx} \|$ can still be large, while conversely, if $\mbox{dim}[\calZ] >  \mbox{dim}[\calU]$ there can now exist a many-to-one mapping from $\{\by,\bz\}$ to $\bx$, in which case the superfluous degrees of freedom can interfere with our ability to disambiguate the role $\by$ plays in predicting $\bx$.  This issue, combined with the fact that we have no mechanism for choosing a suitable $p(\bz)$, implies that deterministic cycle-consistent training is difficult to instantiate.





In contrast, CycleCVAE can effectively circumvent these issues via two key mechanisms that underpin VAE-based models.  First, assume that $\mbox{dim}[\calX] > \mbox{dim}[\calY]$, meaning that $\calX$ represents a higher-dimensional space or manifold relative to $\calY$, consistent with our aforementioned surjective assumption.  Then provided we choose $\mbox{dim}[\calZ]$ sufficiently large, e.g., $\mbox{dim}[\calZ] \geq \mbox{dim}[\calU]$, we would ideally prefer that CVAE regularization somehow prune away the superfluous dimensions of $\bz$ that are not required to produce good reconstructions of $\bx$.\footnote{Note that $\mbox{dim}[\calX]$  refers to the intrinsic dimensionality of $\calX$, which could be a low-dimensional manifold embedded in a higher-dimensional ambient space; same for $\mbox{dim}[\calY]$.}  For example, VAE pruning could potentially be instantiated by setting the posterior distribution of unneeded dimensions of the vector $\bz$ to the prior.  By this we mean that if dimension $j$ is not needed, then $q_\phi(z_j|\bx) = p(z_j)$, uninformative noise that plays no role in improving reconstructions of $\bx$.  This capability has been noted in traditional VAE models \citep{dai2018jmlr,bin2019iclr}, but never rigorously analyzed in the context of CVAE extensions or cycle-consistent training.  In this regard, the analysis in Section \ref{sec:analysis} will elucidate special cases whereby CycleCVAE can provably lead to optimal pruning, the first such analysis of CVAE models, cycle-consistent or otherwise.  This serves to motivate the proposed pipeline as a vehicle for learning an implicit bijection even without knowing the dimensionality or distribution of data from $\calU$, a particularly relevant notion given the difficulty in directly estimating $\mbox{dim}[\calU]$ in practice.

Secondly, because the CycleCVAE model is explicitly predicated upon a \emph{known} prior $p(\bz)$ (as opposed to the unknown distribution of $\bu$), other model components are calibrated accordingly such that there is no need to provide an empirical estimate of an unknown prior.  Consequently, there is no barrier to cycle-consistent training or the generation of new $\bx$ conditioned on $\by$.

\vspace*{-0.2cm}
\section{Formal Analysis of Special Cases} \label{sec:analysis}
\vspace*{-0.2cm}
To the best of our knowledge, there is essentially no existing analysis of cycle-consistent training in the challenging yet realistic scenarios where a bijection between $\bx$ and $\by$ \emph{cannot} be assumed to hold.\footnote{Note that \citep{grover2020alignflow} addresses identifiability issues that arise during cycle training, but only in the context of  strictly bijective scenarios.
}  In this section we present a simplified case whereby the proposed CycleCVAE objective (with added distributional constraints) is guaranteed to have no bad local minimum in a specific sense to be described shortly.  The forthcoming analysis relies on the assumption of a Gaussian CVAE with an affine model for the functions $\{h_\theta,h_\theta^+ \}$; however, the conclusions we draw are likely to be loosely emblematic of behavior in broader regimes of interest.  While admittedly simplistic, the resulting CVAE objective remains non-convex, with a combinatorial number of distinct local minima.  Hence it is still non-trivial to provide any sort of guarantees in terms of associating local minima with `good' solutions, e.g., solutions that recover the desired latent factors, etc.  In fact, prior work has adopted similar affine VAE decoder assumptions, but only in the much simpler case of vanilla VAE models \citep{dai2018jmlr,lucas2019don}, i.e., no cycle training or conditioning as is our focus herein.

\subsection{Affine CycleCVAE Model} \label{sec:affine_cvae}
For analysis purposes, we consider a CVAE model of continuous data  $\bx \in \mathbb{R}^{r_x}$, $\by \in \mathbb{R}^{r_y}$, and $\bz \in  \mathbb{R}^{r_z}$, where $r_x$, $r_y$, and $r_z$ are the respective sizes of $\bx$, $\by$, and $\bz$.  We assume $p(\bz) = \calN(\bz|{\bf 0},\bI)$ and a typical Gaussian decoder $p_{\tiny \theta} \left(\bx | \hat{\by},\bz  \right) = \mathcal{N}(\bx | \bmu_x, \bSigma_x)$, where the mean network satisfies the affine parameterizations
\begin{align} \label{eq:mu_x_definition}
& \bmu_x  =  h_\theta\left(\hat{\by},\bz\right)  =  \bW_{x} \hat{\by} + \bV_{x} \bz + \bb_x, \nonumber 
\\
& \mbox{with}~~ \hat{\by} = h_\theta^+\left(\bx\right) = \bW_y \bx + \bb_y.
\end{align}
In this expression, $\{\bW_x, \bW_y, \bV_x, \bb_x,\bb_y\}$ represents the set of all weight matrices and bias vectors which define the decoder mean $\bmu_x$.  And as is often assumed in practical VAE models, we set $\bSigma_x = \gamma \bI$, where $\gamma > 0$ is a scalar parameter within the parameter set $\theta$.  Despite these affine assumptions, the CVAE energy function can still have a combinatorial number of distinct local minima as mentioned previously.  However, we will closely examine conditions whereby all these local minima are actually global minima that correspond with the optimal inversion of a non-trivial generative
model.

Although we could proceed by allowing the encoder to be arbitrarily complex, when the decoder mean function is forced to be affine and $\bSigma_x = \gamma \bI$, a Gaussian encoder with affine moments is sufficient to achieve the optimal CVAE cost.  Specifically, without any loss of representational capacity, we may choose $q_\phi(\bz | \bx) = \mathcal{N}(\bz | \bmu_z, \bSigma_z)$ with $\bmu_z = \bW_z\bx + \bb_z$ and a diagonal $\bSigma_z = \mbox{diag}[\bs]^2$, where $\bs$ is an arbitrary parameter vector independent of $\bx$.\footnote{Note also that because $\hat{y} = \bW_y \bx + \bb_y$ is an affine function of $\bx$, including this factor in the encoder representation is redundant, i.e., it can be absorbed into $\bmu_z = \bW_z\bx + \bb_z$ without loss of generality.}  Collectively, these specifications lead to the complete parameterization $\theta = \{\bW_x, \bW_y, \bV_x, \bb_x, \gamma \}$, $\phi = \{\bW_z, \bb_z, \bs \}$, and the CVAE energy given by ~ $\ell_x(\theta,\phi) ~ \equiv $
\begin{eqnarray} \label{eq:vae_affine_cost}
\int \Big\{ \mathbb{E}_{q_{\tiny \phi}\left(\bz|\bx \right)} \left[ \tfrac{1}{\gamma} \left\| \left( \bI - \bW_x \bW_y \right)\bx - \bV_x \bz   - \bb_x  \right\|_2^2  \right]  & &  \\
&& \hspace*{-8.4cm} \left. + ~ d \log \gamma  +   \sum_{k=1}^{r_z} \left(s_k^2 - \log s_k^2  \right)  + \left\| \bW_z \bx + \bb_z \right\|_2^2  \Big\} \rho_{gt}^x(d \bx),\right. \nonumber
\end{eqnarray}
noting that, without loss of generality, we have absorbed a $\bW_x\bb_y$ factor into $\bb_x$.

And finally, for the corresponding $\ell_y(\theta)$ model we specify $p_\theta\left(\by|\hat{\bx} \right) = \calN(\by |\bmu_y, \bSigma_y)$ using a shared, cycle-consistent parameterization borrowed from (\ref{eq:mu_x_definition}).  For this purpose, we adopt
\begin{equation}
\bmu_y = h_\theta^+\left( \hat{\bx} \right) = \bW_y \hat{\bx} + \bb_y, 
~~\mbox{with}~~ \hat{\bx} = \bW_{x} \by + \bV_{x} \bz + \bb_x,
\end{equation}
and $\bSigma_y = \gamma \bI$.  Given these assumptions, we have
\begin{equation} \label{eq:y_cycle_cost}
\ell_y(\theta) \equiv  \int \left( \bepsilon_y^{\top} \bSigma_{\epsilon_y}^{-1} \bepsilon_y  + \log \left| \bSigma_{\epsilon_y}  \right| \right)\rho_{gt}^y(d\by)
\end{equation}
excluding irrelevant constants, where $\bepsilon_y \triangleq  \left(\bI - \bW_y \bW_x\right)\by - \bb_y$, $\bSigma_{\epsilon_y}  \triangleq \gamma \bI + \bW_y \bV_x \bV_x^{\top} \bW_y^{\top}$ and again, analogous to before we have absorbed $\bW_y \bb_x$ into $\bb_y$ without loss of generality.

\subsection{Properties of Global/Local Minima}\label{sec:minima}

As a preliminary thought experiment, we can consider the minimization of $\ell_{\tiny \mathrm{cycle}}(\theta,\phi)$, where $\ell_x(\theta,\phi)$ is defined via  (\ref{eq:vae_affine_cost}) and $\ell_y(\theta)$ via (\ref{eq:y_cycle_cost}), but no assumptions are placed on the distributions $\rho_{gt}^x$ and $\rho_{gt}^y$.  In this  situation, it is obvious that even CycleCVAE global minima, were they obtainable, will not generally recover the ground-truth mappings between paired $\bx \sim \rho_{gt}^x$ and $\by \sim \rho_{gt}^y$; there simply will not generally be sufficient capacity.  Furthermore, it can even be shown that under quite broad conditions there will exist a combinatorial number of \emph{non-global} local minima, meaning local minimizers that fail to achieve the lowest possible cost.



Hence we now present a narrower scenario with constraints on the ground-truth data to better align with the affine simplification described in Section \ref{sec:affine_cvae}.   This will allow us to formulate conditions whereby all local minima are actually global minima capable of accurately modeling the ground-truth surjection.  To this end, we define the following affine ground-truth model:

\begin{definition}[Affine Surjective Model]  \label{def:affine_surjective_model}
We define an affine surjective model whereby all matched $\{\bx,\by\}$ pairs satisfy
\begin{equation}
\bx = \bA \by + \bB \bu + \bc    ~~\mbox{and}~~ \by = \bD \bx + \bolde,
\end{equation}
with $\bB \in \operatorname{null}[\bD]$, $\bD \bA  = \bI$, $\bD \bc = -\bolde$, 
$\operatorname{rank}[\bA] = r_y < r_x$ and $\operatorname{rank}[\bB] \leq r_x - r_y$.  Furthermore, we assume that $\by \sim \rho_{gt}^y$ and $\bu \sim \rho_{gt}^u$ are uncorrelated, and the measure assigned to the transformed random variable $\bW \by + \bV \bu$ is equivalent to $\rho_{gt}^y$ iff $\bW = \bI$ and $\bV = {\bf 0}$.  We also enforce that $\by$ and $\bu$ have zero mean and identity covariance, noting that any nonzero mean components can be absorbed into $\bc$.  
Among other things, the stated conditions of the affine surjective model collectively ensure that the mappings $\by \rightarrow \bx$ and $\bx \rightarrow \by$ can be mutually satisfied.  
\end{definition}
Additionally, for later convenience, we also define $r_c \triangleq \operatorname{rank}\left(\mathbb{E}_{\rho_{gt}^x}\left[\bx \bx^\top \right] \right) \leq r_x$.  We then have the following:
\begin{proposition} \label{prop:no_bad_local_min}
Assume that matched pairs $\{\bx,\by\}$ follow the affine surjective model from Definition \ref{def:affine_surjective_model}. Then subject to the constraint $\rho_{gt}^y = \rho_\theta^{\hat{y}}$, where $\rho_\theta^{\hat{y}}$ defines the $\theta$-dependent distribution of $\hat{y}$, all local minima of the CycleVAE objective $\ell_{\tiny \mathrm{cycle}}(\theta,\phi)$, with $\ell_x(\theta,\phi)$ taken from (\ref{eq:vae_affine_cost}) and $\ell_y(\theta)$ from (\ref{eq:y_cycle_cost}), will be global minima in the limit $\gamma \rightarrow 0$ assuming $r_z \geq r_c - r_y$.  Moreover, the globally optimal parameters $\{\theta^*,\phi^* \}$ will satisfy
\begin{eqnarray}
&\bW_x^* = \bA, ~~ \bV_x^* = \left[ \widetilde{\bB}, {\bf 0}\right]\bP, ~~ \bb_x^* = \bc,& \nonumber \\ 
&\bW_y^* = \widetilde{\bD}, ~~ \bb_y^* = -\widetilde{\bD}\bc,&
\end{eqnarray}
where $\widetilde{\bB}$ has $\operatorname{rank}[\bB ]$ columns, $\mbox{span}[\widetilde{\bB}] = \mbox{span}[\bB]$, $\bP$ is a permutation matrix, and $\widetilde{\bD}$ satisfies $\widetilde{\bD}\bA = \bI$ and $\bB \in \mbox{null}[\widetilde{\bD}]$.
\end{proposition}

\noindent Note that in practice, we are free to choose $r_z$ as large as we want, so the requirement that $r_z \geq r_c - r_y$ is not significant.  Additionally, the constraint $\rho_{gt}^y = \rho_\theta^{\hat{y}}$ can be instantiated (at least approximately) by including a penalty on the divergence between these two distributions.  This is feasible using only unpaired samples of $\by$ (for estimating $\rho_{gt}^y$) and $\bx$ (for estimating $\rho_\theta^{\hat{y}}$), and most cycle-consistent training pipelines contain some analogous form of penalty on distributional differences between cycles \citep{DBLP:conf/iclr/LampleCDR18}.  

And finally, if $r_y + \mbox{rank}[\bB] = r_x$, then the dual requirements that $\widetilde{\bD}\bA = \bI$ and $\bB \in \mbox{null}[\widetilde{\bD}]$ will ensure that $\widetilde{\bD}= \bD$ and $\bb_y^* = \bolde$.  However, even if $\widetilde{\bD}\neq \bD$ it is inconsequential for effective recovery of the ground-truth model since any $\bx$ produced by Definition \ref{def:affine_surjective_model} will nonetheless still map to the correct $\by$ when applying $\widetilde{\bD}$ instead of $\bD$.

\begin{figure*}[h]
\begin{minipage}{.63\textwidth}
  \centering
  \includegraphics[width=\linewidth]{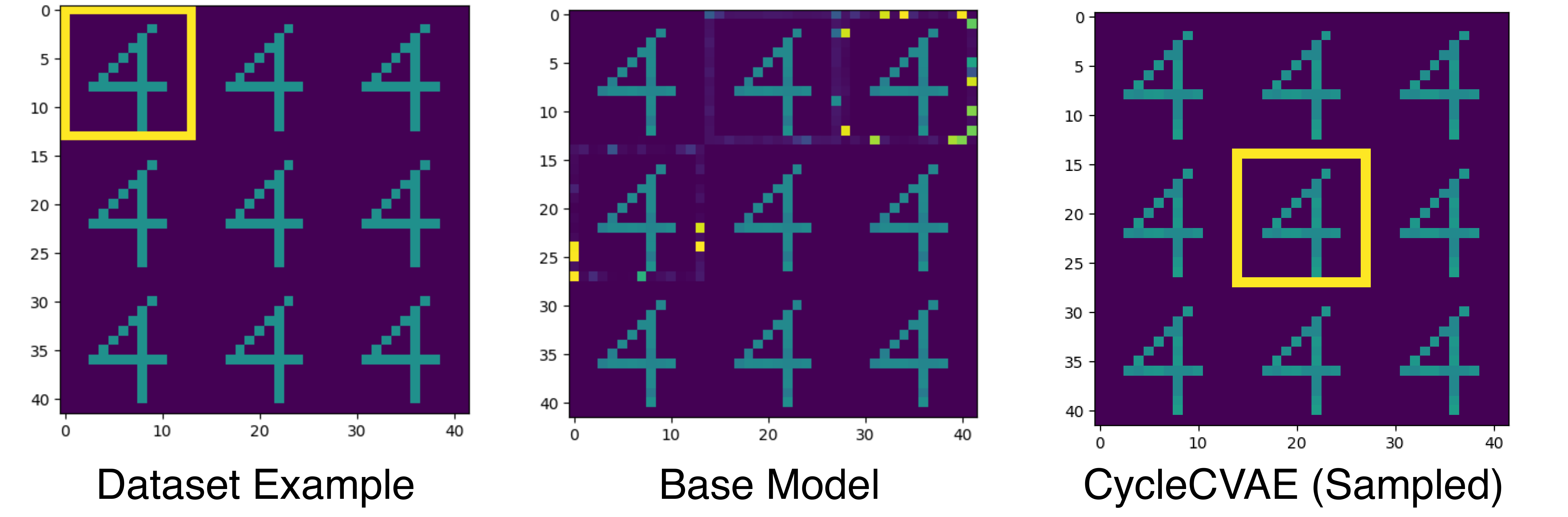}  
  \caption{Left: example image from dataset. Middle: image produced by baseline cycle training with $\by = 4$. Right: a sample image generated by CycleCVAE conditioned on $\by = 4$. For the latter, the position of yellow border is random.  In contrast, the base model fails to learn the random border distribution.}
  \label{fig:toy_res}
\end{minipage}
\hfill
\begin{minipage}{.33\textwidth}
  \centering
  \includegraphics[width=0.9\linewidth]{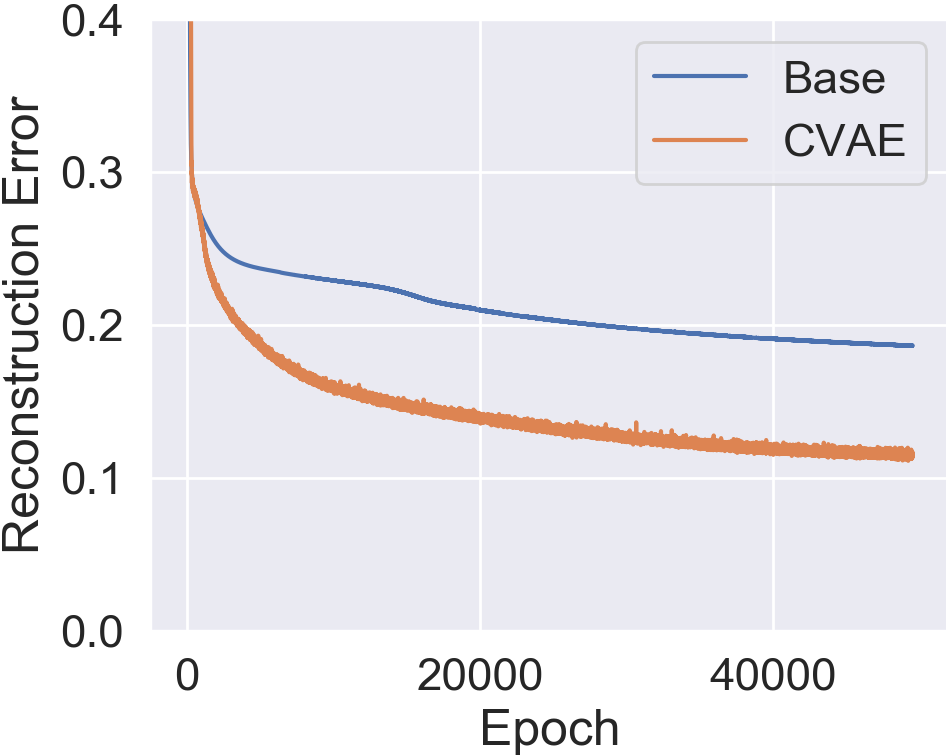}  
  \caption{
  Cycle-consistent reconstruction errors of baseline and CycleCVAE models.
  }
  \label{fig:toy_curve}
\end{minipage}
\end{figure*}

\begin{corollary} \label{cor:implicit_bijection}
Given the same setup as Proposition \ref{prop:no_bad_local_min}, let $\{\bW_z^*,\bb_z^*\}$ denote the CVAE encoder parameters of any minimum.  Then 
$\bW_z^* = \bP \left[ \begin{array}{c} \widetilde{\bW}_z^*  \\ {\bf 0} \end{array} \right]$ and $\bb_z^* = \bP \left[ \begin{array}{c} \widetilde{\bb}_z^*  \\ {\bf 0} \end{array} \right]$, 
where $\widetilde{\bW}_z^*$ has $\operatorname{rank}[\bB ]$ rows, $\widetilde{\bb}_z^* \in \mathbb{R}^{\operatorname{rank}[\bB ]}$, 
and there exists a bijection between $\bx$ and $\{\by,\widetilde{\bmu}_z\}$, with $\widetilde{\bmu}_z \triangleq \widetilde{\bW}_z^*\bx + \widetilde{\bb}_z^*$ (i.e., the nonzero elements of $\bmu_z$).

\end{corollary}

\noindent We will now discuss various take-home messages related to these results.

\subsection{Practical Implications}

As alluded to in Section \ref{sec:cvae_advantages}, we will generally not know in advance $p(\bu)$ or even $\mbox{dim}[\calU]$, which reduces to $r_c - r_y$ in the simplified affine case.  Hence inducing a bijection may seem problematic on the surface.  Fortunately though, Proposition \ref{prop:no_bad_local_min} and Corollary \ref{cor:implicit_bijection} indicate that as long as we choose $r_z \geq \mbox{dim}[\calU]$ in our CVAE model, we can nonetheless still learn an \emph{implicit} bijection between $\bx$ and $\{\by,\widetilde{\bmu}_z\}$, where $\widetilde{\bmu}_z$ are the informative (nonzero) dimensions of $\bmu_z$ that actually contribute to the reconstruction of $\bx$.  In the affine case, these are the dimensions of $\bz$ aligned with nonzero columns of $\widetilde{\bB}$, but in general these dimensions could more loosely refer to the degrees-of-freedom in $\bz$ that, when altered, lead to changes in $\hat{\bx}$.  The remaining superfluous dimensions of $\bz$ are set to the uninformative prior $p(\bz) = \calN(\bz|{\bf 0},\bI)$ and subsequently filtered out by the CVAE decoder module parameterized by $h_\theta(\by,\bz)$.  In this diminutive role, they have no capacity for interfering with any attempts to learn a bijection.

Even so, in non-affine cases it is impossible to guarantee that all local minima correspond with the recovery of $h_{gt}$ and $h_{gt}^+$, or that these functions are even identifiable.  Indeed it is not difficult to produce counterexamples whereby recovery is formally impossible.  However, if $h_{\theta}$ and $h_{\theta}^+$ are chosen with inductive biases reasonably well-aligned with their ground-truth counterparts, up to unidentifiable latent transformations of the unobservable $\bu$, we may expect that an approximate bijection between the true $\bx$ and $\{\by,\widetilde{\bmu}_z\}$ can nonetheless be inferred via cycle-consistent training, at least provided we can avoid suboptimal local minimizers that can be introduced by more complex nonlinear decoder models.


\section{Experiments}\label{sec:experiments}
In this section we first consider a synthetic image experiment that supports the theoretical motivation for \cvaeModel{}.  We then turn to practical real-world evaluations involving the conversion between knowledge graphs and text sequences, a core ingredient of natural language understanding/generation and the application that initially motivated our work.  We then conclude with an enhanced pipeline involving the recent pretrained T5 model \citep{raffel2020exploring}.

\subsection{Synthetic Image Experiment}

We first conduct an experiment designed such that the surjective conditions of Definition \ref{def:affine_surjective_model} in Section~\ref{sec:minima} are loosely satisfied (see supplementary for reasons). As shown in Figure~\ref{fig:toy_res} (left panel), each data sample has three components: a digit, its image, and a decorative yellow border.  The digit takes value in $\{0,\ldots,9\}$ and is represented by a 10-dim one-hot vector $\by$. The corresponding image $\bx$ involves $3 \times 3$ tiles, and each tile contains the same image of digit $\by$. One of the 9 tiles is decorated with a 1-pixel-wide yellow border, and which tile will have this border is determined by $\bu$, a 9-dim one-hot vector indicating the 9 possible tiles.

We train two models on this dataset, a base model using standard cycle training, and our CycleCVAE that incorporates the proposed CVAE into a baseline cycle model (see supplementary for network description and details).
After training,  generated samples of the two approaches when presented with the digit `4' are shown in Figure~\ref{fig:toy_res} (middle and right panels). The base model fails to learn the yellow border as it cannot handle the one-to-many mapping from digits to images.  Meanwhile the random CycleCVAE sample correctly places the border around one of the tiles (different CycleCVAE samples simply move the border to different tiles as desired; see supplementary).  Finally, consistent with these generation results, the training curves from Figure~\ref{fig:toy_curve} reveal that the reconstruction error of the base model, which assumes a bijection, plateaus at a significantly higher value than the CycleCVAE model.


\subsection{Knowledge Graph to Text Conversion}\label{sec:webnlg}

We now turn to more challenging real-world experiments involving the surjective mapping between knowledge graphs and text sequences.  Here the ideal goal is to generate diverse, natural text from a fixed knowledge graph, or extract the knowledge graph from a piece of text.  To this end we compare CycleCVAE against SOTA methods on the widely-used WebNLG graph-to-text dataset \citep{DBLP:conf/inlg/GardentSNP17}.


\paragraph{WebNLG Dataset and Test Setup}  WebNLG data is extracted from DBPedia, where each graph consists of 2--7 nodes and the corresponding text is descriptions of these graphs collected by crowd-sourcing. 
We follow the preprocessing of \citep{DBLP:conf/naacl/MoryossefGD19} and obtain 13K training, 1.6K validation, and 5K test text-graph pairs.  Please see the supplementary for details of the CycleCVAE architecture explicitly designed for handling text and graph data.  Note that we did not include any additional penalty function on the divergence between $\rho_{gt}^y$ and $\rho_\theta^{\hat{y}}$; the architecture inductive biases were sufficient for good performance.


\paragraph{Metrics}
We measure performance using three metrics: (1) text generation quality with the standard BLEU score \citep{papineni2002bleu},\footnote{BLEU (\%) counts the 1- to 4-gram overlap between the generated sentence and ground truth.} (2) graph construction accuracy via the F1 score of the edge predictions among given entity nodes, and (3) text diversity. Text diversity is an increasingly important criterion for NLP because the same meaning can be conveyed in various expressions, and intelligent assistants should master such variations.
We evaluate diversity by reporting
the number of distinct sentence variations obtained after running the generation model 10 times. 



\paragraph{Accuracy Results}
Since cycle training only requires unsupervised data, we have to break the text-graph pairs to evaluate unsupervised performance. In this regard, there are two ways to process the data.  First, we can use 100\% of the training data and just shuffle the text and graphs so that the matching/supervision is lost.  This is the setting in Table~\ref{tab:webnlg100}, which allows for direct head-to-head comparisons with SOTA supervised methods (assuming no outside training data). The supervised graph-to-text baselines include \textit{Melbourne}
(introduced in \cite{DBLP:conf/inlg/GardentSNP17}), \textit{StrongNeural}, \textit{BestPlan} \citep{DBLP:conf/naacl/MoryossefGD19}, \textit{\segAlign{}} \citep{shen2020neural}, and \textit{G2T} \citep{DBLP:conf/naacl/Koncel-Kedziorski19}.
Supervised text-to-graph models include \textit{OnePass} \citep{DBLP:conf/acl/WangTYCWXGP19}, and \textit{T2G}, a BiLSTM model we implemented. Unsupervised methods include \textit{RuleBased} and \textit{GT-BT} both by \citep{schmitt2019unsupervised}.
Finally, \textit{\baseModel{}} is our deterministic cycle training model with the architectural components borrowed from CycleCVAE. Notably, from Table~\ref{tab:webnlg100} we observe that our model outperforms other unsupervised methods, and it is even competitive with SOTA supervised models in both the graph-to-text (BLEU) and text-to-graph (F1) directions.


\begin{table}[ht]
    \centering \small
    \resizebox{\columnwidth}{!}{%
    \begin{tabular}{lccc}
        \toprule
        & {Text(BLEU)} & Graph(F1) & \#Variations \\
        \hline
        \multicolumn{4}{l}{\textbf{{Supervised (100\%)}}}
        \\
        \quad {Melbourne} & 45.0 & -- & 1 \\
        \quad {StrongNeural
        } & 46.5 & -- & 1 \\
        \quad {BestPlan
        } & 47.4 & -- & 1 \\
        \quad {\segAlign{}} & 46.1 & -- & 1
        \\
        \quad {G2T
        } & 45.8 & -- & 1 \\
        \quad {OnePass
        } & -- & 66.2 & --
        \\
        \quad {T2G} & -- & 60.6 & -- \\
        \multicolumn{4}{l}{\textbf{{Unsupervised (100\%, Shuffled)}}} \\
        \quad {RuleBased} & 18.3 & 0 & 1
        \\
        \quad {GT-BT} & 37.7 & 39.1 & 1
        \\
        \quad \baseModel{} (Ours) & 46.2 & 61.2 & 1 \\
        \quad {\cvaeModel{}} (Ours) & 46.5 & 62.6 & 4.67 \\
        \bottomrule
    \end{tabular}
    }
    \caption{Performance on the full WebNLG dataset.}
    \label{tab:webnlg100}
\end{table}
\begin{table}[ht]
    \centering \small
    \resizebox{\columnwidth}{!}{%
    \begin{tabular}{lccc}
        \toprule
        & Text(BLEU) & Graph(F1) & \#Variations \\
        \hline
        \multicolumn{4}{l}{\textbf{{Supervised (50\%)}}} \\
        \quad {G2T} & 44.5 & -- & 1 \\
        \quad {T2G} & -- & 59.7 & --  \\
        \multicolumn{4}{l}{\textbf{{Unsupervised (first 50\% text, last 50\% graph)}}} \\
        \quad \baseModel{} (Ours) & 43.1 & 59.8 & 1 \\
        \quad {\cvaeModel{}} (Ours) & 43.3 & 60.0 & 4.01 \\
        \bottomrule
    \end{tabular}
    }
    \caption{Performance on WebNLG with 50\% data.}
    \label{tab:webnlg50}
\end{table}

In contrast, a second, stricter unsupervised protocol involves splitting the dataset into two halves, extracting text from the first half, and graphs from the second half.  This is the setting in Table~\ref{tab:webnlg50}, which avoids the possibility of seeing any overlapping entities during training.  Although performance is slightly worse given less training data, the basic trends are the same.

\paragraph{Diversity Results} From Tables~\ref{tab:webnlg100} and \ref{tab:webnlg50} we also note that CycleCVAE can generate on average more than 4 different sentence types for a given knowledge graph; all other SOTA methods can only generate a single sentence per graph.  Additionally, we have calculated that CycleCVAE generates more than two textual paraphrases for 99\% of test instances, and the average edit distance between two paraphrases is 12.24 words (see supplementary). Moreover, CycleCVAE text diversity does not harm fluency and semantic relevance as the BLEU score is competitive with SOTA methods as mentioned previously. 
\begin{table}[ht]
    \centering
    \small
    \begin{tabular}{p{7.8cm}}
    \toprule
    \multicolumn{1}{c}{Diverse Text Output Generated by \cvaeModel{}} \\ \hline
         -- The population density of Arlington, Texas is 1472.0. \\
         -- Arlington, Texas has a population density of 1472.0. \\ \hline
         -- Alan Bean, who was born in Wheeler, Texas, is now ``retired.'' \\
         --~Alan Bean is a United States citizen who was born in Wheeler, Texas. He is now ``retired.'' \\
    \bottomrule
    \end{tabular}
    \caption{Every two variations are generated by \cvaeModel{} from the same knowledge graph.}
    \label{tab:cvae_ex}
\end{table}

We list text examples generated by our model in Table~\ref{tab:cvae_ex}, with more in the supplementary. The diverse generation is a significant advantage for many real applications. For example, it can make automated conversations less boring and simulate different scenarios. And diversity can push model generated samples closer to the real data distribution because there exist different ways to verbalize the same knowledge graph (although diversity will not in general improve BLEU scores, and can sometimes actually lower them).

\subsection{Integrating CycleCVAE with T5}
Previous graph-text results are all predicated on no outside training data beyond WebNLG.  However, we now consider an alternative testing scenario whereby outside training data can be incorporated by integrating CycleCVAE with a large pretrained T5 sequence-to-sequence model \citep{raffel2020exploring}.  Such models have revolutionized many NLP tasks and can potentially improve the quality of the graph-to-text direction in cycle training on WebNLG.  To this end, we trained a CycleCVAE model, with the function $h_\theta(\by,\bz)$ formed from a pretrained T5 architecture (see supplementary for details).  Results are shown in Table~\ref{tab:t5}, where unsupervised CycleCVAE+T5 produces a competitive BLEU score relative to fully supervised T5 baselines.  It also maintains diversity of generated text sequences. 


\begin{table}[ht]
    \centering
    \small
    \begin{tabular}{lcc}
    \toprule
        {Model} & {BLEU} & {\#Vars.} \\ \hline
        Supervised~T5 \citep{kale2020text} & 57.1 & 1 \\
        Supervised~T5 \citep{ribeiro2020investigating} & 57.4 & 1 \\
        Supervised~T5 (Our Impl.) & 56.4 & 1\\
        Unsupervised~\cvaeModel{}+T5 & 55.7 & 3.84 \\
    \bottomrule
    \end{tabular}
    \caption{Text generation results with T5 on WebNLG.}
    \label{tab:t5}
\end{table}

\section{Conclusion}
We have proposed CycleCVAE for explicitly handling non-bijective surjections commonly encountered in real-world applications of unsupervised cycle-consistent training. Our framework has both a solid theoretical foundation and strong empirical performance on practical knowledge graph-to-text conversion problems.  For future work we can consider extending CycleCVAE to handle many-to-many (non-bijective, non-surjective) mappings, or unsolved applications such as conversions between scene graphs and realistic images (which remains extremely difficult even with supervision). 



\printbibliography

\onecolumn

\aistatstitle{Supplementary Materials}

The supplementary file includes additional content related to the following:
\begin{enumerate}
    \item Synthetic Experiments (Section~5.1 in main paper):  We explain why the synthetic data loosely align with Definition 1, describe the network architectures of the \baseModel{} and \cvaeModel{} models, and include additional generation results.
    \item WebNLG Experiments (Section~5.2 in main paper): We describe the experimental setup for WebNLG, including the task description, cycle-consistency model design, and all baseline and implementation details.  We also include an ablation study varying $\mbox{dim}[\bz]$.
    \item T5 Extension (Section~5.3 in main paper): We provide details of the \cvaeModel{}+T5 extension and include additional generated samples showing textual diversity.
    \item Proof of Proposition 2.
    \item Proof of Corollary 3.
    
\end{enumerate}

\section{Synthetic Dataset Experimental Details and Additional Results}
\subsection{Dataset Description and Relation to Definition 1}
To motivate how the synthetic data used in Section 5.1 from the main paper at least partially align with Definition
1, we let $\bc$ and $\bolde$ be zero vectors and $\bA \in \mathbb{R}^{d\times10}$ be a $d\times10$ transformation matrix from images to digits, where $d$ is the total number of pixels in each image $\bx$. In other words, each column $i \in \{0,1,\dots,9\}$ of $\bA$ is a linearized pixel sequence of the 2D image of digit $i$ from top left to bottom right. Based on $\bA$, we construct an example inverse matrix $\bD$ so that $\bD\bA=\bI$. Specifically, $\bD$ can be a $10\times d$ matrix where each row $i \in \{0,1,\dots,9\}$ is a linearized pixel sequence of a masked version of the image of the digit $i$, and this image can have, for example, only one non-zero pixel that is sufficient to distinguish the digit $i$ from all other nine possibilities. 
We also construct $\bB$, a $d\times9$ transformation matrix from the image to the border position, which surrounds one out of the nine  tiles in each image. Each column $i \in \{0,1,\dots,8\}$ of $\bB$ is a linearized pixel sequence of the 2D image of the border surrounding the $i$-th tile. 
Since the patterns of the digit and border do not share any non-zero pixels, we should have that $\bD\bB={\bf 0}$. Moreover, each digit's image is distinct and cannot be produced by combining other digit images, so $\operatorname{rank}[\bA]=r_y$ and also $r_y\leq r_x$ because border patterns are orthogonal to digit patterns. Hence, we also have $\operatorname{rank}[\bB]\leq r_x - r_y$. Note however that the synthetic data do not guarantee that $\bW\by+\bV\bu$ is equivalent to $\rho_{gt}^y$ iff $\bW = \bI$ and $\bV = {\bf 0}$. 

\subsection{Network Architectures}
We train two models on this dataset, a base model CycleBase using standard cycle training, and our \cvaeModel{} that incorporates the proposed CVAE into a baseline cycle model.
\paragraph{\baseModel{}}
The base model uses multilayer perceptrons (MLPs) for both the image($\bx$)-to-digit($\by$) mapping $h_\theta^+(\bx)$ (shared with CycleCVAE), and the digit($\by$)-to-image($\bx$) mapping denoted $h_\theta^{\mathrm{Base}}(\by)$. Each MLP hidden layer (two total) has 50 units with the $\mathrm{tanh}$ activation function. The last layer of $h_\theta^+(\bx)$ uses a softmax function to output a vector of probabilities $\balpha$ over the ten digits, and therefore we can apply $p_\theta(\by | \bx) = \mathrm{Cat}(\by|\balpha)$, a categorical distribution conditioned on $\balpha$, for training purposes. The last layer of digit-to-image $h_\theta^{\mathrm{Base}}(\by)$ adopts a per-pixel sigmoid function (since the value of each pixel is between $0$ and $1$), and we assume $p_\theta(\bx|\by)$ is based on the binary cross entropy loss.

\paragraph{\cvaeModel{}} 
Our \cvaeModel{} uses the same function $h_\theta^+(\bx)$ as the base model.  However, for the digit-to-image generation direction, \cvaeModel{} includes a 1-dimensional latent variable $\bz$ sampled from $\mathcal{N}(\bmu_x, \bSigma_x)$, where $\bmu_x$ and $\bSigma_x$ are both learned by 50-dimensional, 3-layer MLPs (including output layer) with input $\bx$. Then $h_\theta(\by, \bz)$ takes the digit $\by$ and latent variable $\bz$ as inputs to another 3-layer MLP with 50 hidden units and the same activation function as the base model.

\subsection{Generation Results}
In addition to Figure 1 in the main paper, we list more example images generated by our model in the figure below. As we can see, the base model fails to learn the diverse border which should randomly surround only one of the nine tiles. However, \cvaeModel{} learns the border in its latent variable $\bz$ and by random sampling, \cvaeModel{} can generate an arbitrary border around one of the nine digits as expected.
\begin{figure}[ht]
    \centering
    \includegraphics[width=\textwidth]{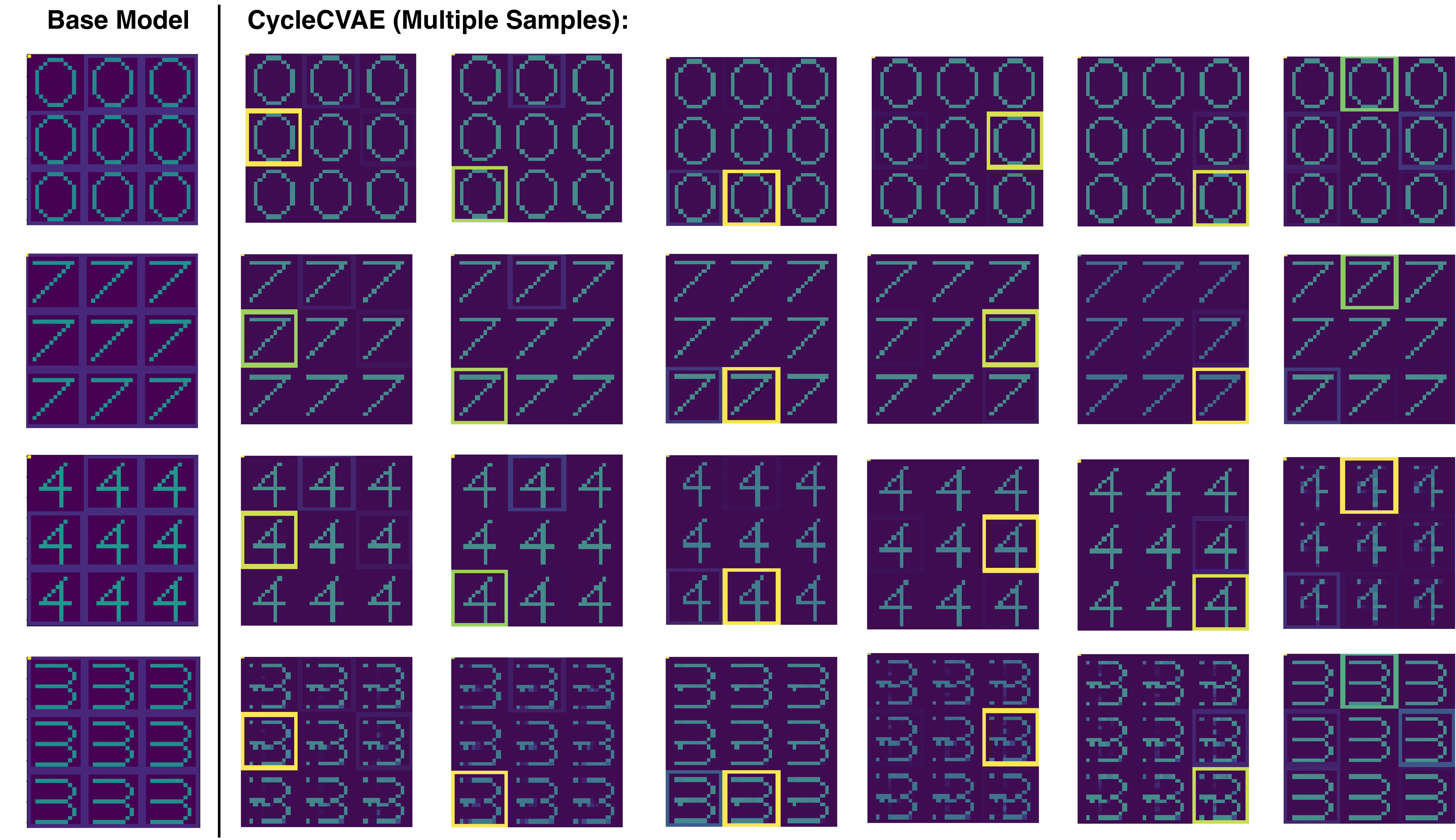}
    \caption{Example images generated by \cvaeModel{}.}
    \label{fig:toy_gen}
\end{figure}

\section{WebNLG Experimental Setup and Ablation Study}

The WebNLG dataset\footnote{It can be downloaded from \url{https://webnlg-challenge.loria.fr/challenge_2017/}.} is widely used for conversions between graph and text. Note that WebNLG is the most appropriate dataset for our purposes because in other candidates (e.g., relation extraction datasets \citep{walker2006ace}) the graphs only contain a very small subset of the information in the text. 

\subsection{Task Description}
\begin{figure}
    \centering
    \includegraphics[width=\textwidth]{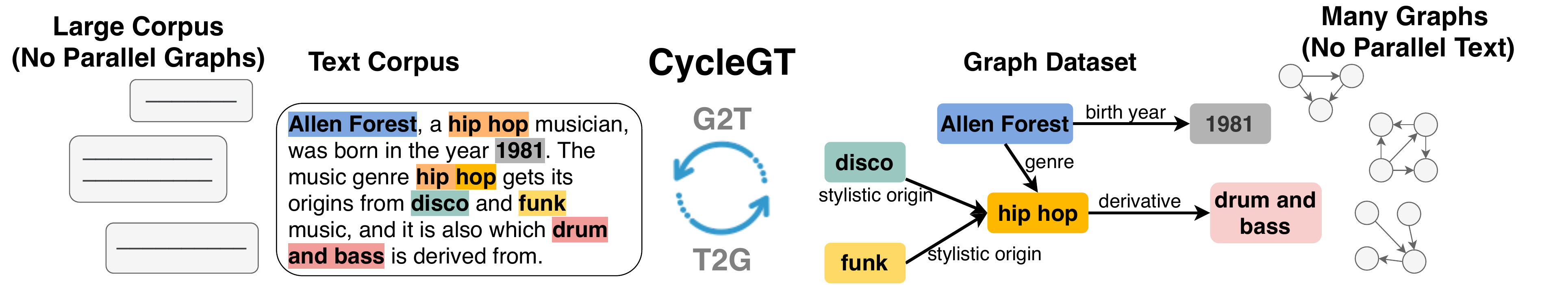}
    \caption{The graph-to-text generation task aims to verbalize a knowledge graph, while the text-to-graph task extracts the information of text into the form of a knowledge graph.}
    \label{fig:kg_to_txt}
\end{figure}

The WebNLG experiment includes two directions: text-to-graph (T2G) and graph-to-text (G2T) generation. The G2T task aims to produce descriptive text that verbalizes the graphical data. For example, the knowledge graph triplets ``\textit{(Allen Forest, genre, hip hop), (Allen Forest, birth year, 1981)}'' can be verbalized as ``\textit{Allen Forest, a hip hop musician, was born in 1981}.'' This has wide real-world applications, for instance, when a digital assistant needs to translate some structured information (e.g., the properties of the restaurant) to the human user. 
The other task, T2G is also important, as it extracts structures in the form of knowledge graphs from the text, so that all entities become nodes, and the relationships among entities form edges. It can help many downstream tasks, such as information retrieval and reasoning. The two tasks can be seen as a dual problem, as shown in Figure~\ref{fig:kg_to_txt}.

Specifically, for unsupervised graph-to-text and text-to-graph generation, we have two non-parallel datasets:
\begin{itemize}[leftmargin=15px]
    \item A text corpus $\bX = \{\bx_i\}^N_{i=1}$ consisting of $N$ text sequences, and
    \item A graph dataset $ \bY = \{\by_j\}^M_{j=1}$ consisting of $M$ graphs.
\end{itemize}
The constraint is that the graphs and text contain the same distribution of latent content, but are different forms of surface realizations, i.e., there is no alignment providing matched pairs.  Our goal is to train two models in an unsupervised manner: $h_\theta $ that generates text based on the graph, and $h_\theta^+$ that produces a graph based on text.

\subsection{Cycle Training Models}
\paragraph{\baseModel{}}
Similar to the synthetic experiments mentioned above, we first propose the base cycle training model \baseModel{} that jointly learns graph-to-text and text-to-graph generation. To be consistent with our main paper, we denote text as $\bx$ and graphs as $\by$, and the graph-to-text generation is a one-to-many mapping. The graph cycle, $\by\rightarrow \hat \bx \rightarrow \hat \by$ is as follows: Given a graph $\by$, the cycle-consistent training first generates synthetic text $\hat \bx = h_\theta^\mathrm{Base} (\by)$, and then uses it to reconstruct the original graph $\hat \by = h_\theta^+(\hat \bx)$. The loss function is imposed to align the generated graph $\hat \by$ with the original graph $\by$. Similarly, the text cycle, $\bx\rightarrow \hat \by \rightarrow \hat \bx$, is to align $\bx$ and the generated $\hat \bx$. Both loss functions adopt the cross entropy loss.

Specifically, we instantiate the graph-to-text module $h_\theta^\mathrm{Base} (\by)$ with the GAT-LSTM model proposed by \citep{DBLP:conf/naacl/Koncel-Kedziorski19}, and the text-to-graph module $h_\theta^+(\bx)$ with a simple BiLSTM model we implemented. The GAT-LSTM module has two layers of graph attention networks (GATs) with 512 hidden units, and two layers of a LSTM text decoder with multi-head attention over the graph node embeddings produced by GAT. This attention mechanism uses four attention heads, each with 128 dimensions for self-attention and 128 dimension for cross-attention between the decoder and node features. The BiLSTM for text-to-graph construction uses 2-layer bidirectional LSTMs with 512 hidden units. 

\paragraph{\cvaeModel{}}
Our \cvaeModel{} uses the same  $h_\theta^+(\bx)$ as the base model. As for $h_\theta(\by, \bz)$ (the CycleCVAE extension of CycleBase), we first generate a 10-dimensional latent variable $\bz$ sampled from $q_\phi(\bz|\bx) = \mathcal{N}(\bmu_x, \bSigma_x)$, where $\bmu_x$ and $\bSigma_x$ are both learned by bidirectional LSTMs plus a fully connected feedforward layer. We form  $p(\bz | \by)$ as a Gaussian distribution whose mean and variance are learned from a fully connected feedforward layer which takes in the feature of the root node of the GAT to represent the graph. Note that applying this $p(\bz | \by)$ as the CycleCVAE prior is functionally equivalent to using a more complicated encoder, as mentioned in the main paper.

\paragraph{Implementation Details}
For both cycle models, we adopt the Adam optimizer with a learning rate of $5\mathrm{e}{-5}$ for the text-to-graph modules, and learning rate of $2\mathrm{e}{-4}$ for graph-to-text modules.
For the graph-to-text module, we re-implement the GAT-LSTM model \citep{DBLP:conf/naacl/Koncel-Kedziorski19} using the DGL library \citep{DBLP:journals/corr/abs-1909-01315}. 
Our code is available \codeURL{}.

\subsection{Details of Competing Methods}
\paragraph{Unsupervised Baselines}
As cycle training models are unsupervised learning methods, we first compare with unsupervised baselines. \textit{RuleBased} is a heuristic baseline proposed by \citep{schmitt2019unsupervised} which simply iterates through the graph and concatenates the text of each triplet. For example, the triplet ``(AlanShepard, occupation, TestPilot)'' will be verbalized as ``Alan Shepard occupation test pilot.'' If there are multiple triplets, their text expressions will be concatenated by ``and.'' The other baseline, \textit{UMT} \citep{schmitt2019unsupervised}, formulates the graph and text conversion as a sequence-to-sequence task and applies a standard unsupervised machine translation (UMT) approach. It serializes each triplet of the graph in the same way as RuleBased, and concatenates the serialization of all triplets in a random order, using special symbols as separators. 

\paragraph{Supervised Baselines} We also compare with \textit{supervised} systems using the original supervised training data. Since there is no existing work that jointly learns graph-to-text and text-to-graph in a supervised way, we can only use models that address one of the two tasks. For graph-to-text generation, we list the performance of state-of-the-art supervised models including (1) \textit{Melbourne}, the best supervised system submitted to the WebNLG challenge 2017 \citep{DBLP:conf/inlg/GardentSNP17}, which uses an encoder-decoder architecture with attention, (2) \textit{StrongNeural} \citep{DBLP:conf/naacl/MoryossefGD19} which improves the common encoder-decoder model, (3) \textit{BestPlan} \citep{DBLP:conf/naacl/MoryossefGD19} which uses a special entity ordering algorithm before neural text generation, (4) \textit{G2T} \citep{DBLP:conf/naacl/Koncel-Kedziorski19} which is the same as the GAT-LSTM architecture adopted in our cycle training models,
and (5) \textit{\segAlign{}} \citep{shen2020neural}, which segments the text into small units, and learns the alignment between data and target text segments. The generation process uses the attention mechanism over the corresponding data piece to generate the corresponding text. 
For text-to-graph generation, we compare with state-of-the-art models including \textit{OnePass} \citep{DBLP:conf/acl/WangTYCWXGP19}, a BERT-based relation extraction model, and \textit{T2G}, the BiLSTM model that we adopt as the text-to-graph component in the cycle training of \baseModel{} and \cvaeModel{}.

\subsection{Ablation Study}
We conduct an ablation study using the 50\%:50\% unsupervised data of WebNLG. Note that our models do not use an adversarial term, so we only tune the CVAE latent dimension to test robustness to this factor. The hyperparameter tuning of the size of the latent dimension is shown in Table~\ref{tab:tune}, where we observe that our \cvaeModel{} is robust against different $\bz$ dimensions.  Note that because $\bz$ is continuous while generated text is discrete, just a single dimension turns out to be adequate for good performance for these experiments.  Even so, the encoder variance can be turned up to avoid `overusing' any continuous latent dimension to roughly maintain a bijection.
\begin{table}[ht]
    \centering
    \begin{tabular}{lccc}
    \toprule
         &  & Text (BLEU) & Diversity (\# Variations) \\ \hline
        Latent Dimension \\
        \quad $\mbox{\bz}=1$ & & 46.3 & 4.62\\
        \quad $\mbox{\bz}=10$ & & 46.5 & 4.67 \\
        \quad $\mbox{\bz}=50$ & & 46.2 & 4.65\\
    \bottomrule
    \end{tabular}
    \caption{Text quality (by BLEU scores) and diversity (by the number of variations) under different dimensions of $\bz$.}
    \label{tab:tune}
\end{table}

\section{T5 Model Details and More Generated Samples}
\subsection{\cvaeModel{}+T5 Implementational Details}


We adopted the pretrained T5 model \citep{raffel2020exploring} to replace the GAT-LSTM architecture that we previously used for the graph-to-text module within the cycle training. T5 is a sequence-to-sequence model that takes as input a serialized graph (see the serialization practice in \cite{schmitt2019unsupervised,ribeiro2020investigating,kale2020text}) and generates a text sequence accordingly. We finetune the T5 during training with the Adam optimizer using a learning rate of $5\mathrm{e}{-5}$.

\subsection{Additional Text Diversity Examples}
We list the text diversity examples generated by \cvaeModel{}+T5 in Table~\ref{tab:txt_diversity}.
\begin{table}[H]
    \centering
\begin{tabular}{c p{15.8cm}}
\toprule
No. & Variations \\ \hline
     
     \multirow{4}{*}{1} & -- Batagor, a variation of Siomay and Shumai, can be found in Indonesia, where the leader is Joko Widodo and Peanut sauce is an ingredient. \\
     & -- Batagor is a dish from Indonesia, where the leader is Joko Widodo and the main ingredient is Peanut sauce. It can also be served as a variation of Shumai and Siomay. \\ 
     \hline
     \multirow{4}{*}{2} & -- The AMC Matador, also known as ``American Motors Matador'', is a Mid-size car with an AMC V8 engine and is assembled in Thames, New Zealand.  \\
     & -- AMC Matador, also known as ``American Motors Matador'', is a Mid-size car. It is made in Thames, New Zealand and has an AMC V8 engine. \\
     \hline
     \multirow{3}{*}{3} & -- Aleksandr Chumakov was born in Moscow and died in Russia. The leader of Moscow is Sergey Sobyanin. \\ 
     & -- Aleksandr Chumakov, who was born in Moscow, was a leader in Moscow where Sergey Sobyanin is a leader. He died in Russia.\\
     \hline
     \multirow{4}{*}{4} & -- A Wizard of Mars is written in English language spoken in Great Britain. It was published in the United States, where Barack Obama is the president. \\
     & -- A Wizard of Mars comes from the United States where Barack Obama is the leader and English language spoken in Great Britain.\\
     \hline
     \multirow{3}{*}{5} & -- The Addiction (journal), abbreviated to ``Addiction'', has the ISSN number ``1360-0443'' and is part of the academic discipline of Addiction. \\
     & -- Addiction (journal), abbreviated to ``Addiction'', has the ISSN number ``1360-0443''. \\
     \hline
     \multirow{3}{*}{6} & -- Atlantic City, New Jersey is part of Atlantic County, New Jersey Atlantic County, New Jersey, in the United States. \\
     & -- Atlantic City, New Jersey is part of Atlantic County, New Jersey, United States. \\
     \hline
     \multirow{4}{*}{7} & -- Albuquerque, New Mexico, United States, is lead by the New Mexico Senate, led by John Sanchez and Asian Americans. \\
     & -- Albuquerque, New Mexico, in the United States, is lead by the New Mexico Senate, where John Sanchez is a leader and Asian Americans are an ethnic group. \\
     \hline
     \multirow{4}{*}{8} & -- Aaron Turner plays the Electric guitar and plays Black metal, Death metal and Black metal. He also plays in the Twilight (band) and Old Man Gloom. \\
     & -- Aaron Turner plays the Electric guitar and plays Black metal. He is associated with the Twilight (band) and Old Man Gloom. He also plays Death metal. \\
     \bottomrule
     
\end{tabular}
    \caption{Examples of diverse text generated by \cvaeModel{} based on the same input knowledge graph.}
    \label{tab:txt_diversity}
\end{table}

\section{Proof of Proposition \ref{prop:no_bad_local_min}}
The high-level proof proceeds in several steps.  First we consider optimization of $\ell_x(\theta,\phi)$ over $\phi$ to show that no suboptimal local minima need be encountered.  We then separately consider optimizing $\ell_x(\theta,\phi)$ and $\ell_y(\theta)$ over the subset of $\theta$ unique to each respective loss.  Next we consider jointly optimizing the remaining parameters residing between both terms.  After assimilating the results, we arrive at the stated result of Proposition \ref{prop:no_bad_local_min}.  Note that with some abuse of notation, we reuse several loss function names to simplify the exposition; however, the meaning should be clear from context.

\subsection{Optimization over encoder parameters $\phi$ in $\ell_x(\theta,\phi)$}

The energy term from the $\bx \rightarrow \hat{\by} \rightarrow \hat{\bx}$ cycle can be modified as
\begin{eqnarray} \label{eq:vae_cost_tot_affine_sup}
\ell_x(\theta,\phi) & = &
\int \bigg\{ \mathbb{E}_{q_{\tiny \phi}\left(\bz|\bx \right)} \left[ \tfrac{1}{\gamma} \left\| \bx - \bmu_x  \right\|_2^2  \right]  +  d \log \gamma  +   \sum_{k=1}^{r_z} \left(s_k^2 - \log s_k^2  \right)  + \left\| \bmu_z \right\|_2^2  \bigg\} \rho_{gt}^x(d \bx) \nonumber \\
& = & \int \bigg\{ \mathbb{E}_{q_{\tiny \phi}\left(\bz|\bx \right)} \left[ \tfrac{1}{\gamma} \left\| \left( \bI - \bW_x \bW_y \right)\bx - \bV_x \bz  - \bW_x \bb_y - \bb_x  \right\|_2^2  \right]  \nonumber \\
&& \hspace*{-0.0cm}  + ~ d \log \gamma  +   \sum_{k=1}^{r_z} \left(s_k^2 - \log s_k^2  \right)  + \left\| \bW_z \bx + \bb_z \right\|_2^2  \bigg\} \rho_{gt}^x(d \bx)  \\
& = & \int  \bigg\{ \tfrac{1}{\gamma} \left\|  \left( \bI - \bW_x \bW_y \right)\bx - \bV_x \left( \bW_z \bx + \bb_z \right)   - \bW_x \bb_y - \bb_x  \right\|_2^2  \nonumber \\
& & + ~ d \log \gamma  +    \sum_{k=1}^\kappa \left(s_k^2 - \log s_k^2  + \tfrac{1}{\gamma}s_k^2 \| \bv_{x,k}\|_2^2 \right) + \| \bW_z \bx + \bb_z \|_2^2  \bigg\} \rho_{gt}^x(d \bx), \nonumber
\end{eqnarray}
where $\bv_{x,k}$ denotes the $k$-th column of $\bV_x$.  Although this expression is non-convex in each $s_k^2$, by taking derivatives and setting them equal to zero, it is easily shown that there is a single stationary point that operates as the unique minimum.  Achieving the optimum requires only that $s_k^2 = \left[\tfrac{1}{\gamma}\| \bv_{x,k}\|_2^2 + 1 \right]^{-1}$ for all $k$.  Plugging this value into (\ref{eq:vae_cost_tot_affine_sup}) then leads to the revised objective
\begin{eqnarray} \label{eq:vae_cost_tot_affine2_sup}
\ell_x(\theta,\phi) &  \equiv  & \int \bigg\{ \tfrac{1}{\gamma} \left\|  \left( \bI - \bW_x \bW_y \right)\bx - \bV_x \left( \bW_z \bx + \bb_z \right)   - \bW_x \bb_y - \bb_x   \right\|_2^2   \\
&&  +~~ \sum_{k=1}^\kappa \log \left( \tfrac{1}{\gamma} \| \bv_{x,k}\|_2^2 + 1 \right) + d \log \gamma  + \| \bW_z \bx + \bb_z \|_2^2  \bigg\} \rho_{gt}^x(d \bx)  \nonumber
\end{eqnarray}
ignoring constant terms.  Similarly we can optimize over $\bmu_z = \bW_z \bx + \bb_z$ in terms of the other variables.  This is just a convex, ridge regression problem, with the optimum uniquely satisfying
\begin{equation} \label{eq:mu_z_opt}
\bW_z \bx + \bb_z = \bV_x^{\top} \left( \gamma \bI + \bV_x\bV_x^{\top} \right)^{-1} \left[\left( \bI - \bW_x \bW_y \right)\bx - \bW_x \bb_y -\bb_x \right],
\end{equation}
which is naturally an affine function of $\bx$ as required.  After plugging (\ref{eq:mu_z_opt}) into (\ref{eq:vae_cost_tot_affine2_sup}), defining
$\bepsilon_x \triangleq \left( \bI - \bW_x \bW_y \right)\bx - \bW_x \bb_y - \bb_x$, and applying some linear algebra manipulations, we arrive at
\begin{eqnarray} \label{eq:vae_cost_tot_affine3_sup}
\bar{\ell}_x(\theta)  &  \triangleq  &  \min_{\phi} ~ \ell_x(\theta,\phi)  \\
&  =  &  \int \left\{ \bepsilon_x^{\top} \left( \bV_x \bV_x^{\top}  + \gamma \bI\right)^{-1}\bepsilon_x \right\}  \rho_{gt}^x(d \bx)  +  \sum_{k=1}^\kappa \log \left( \| \bv_{x,k}\|_2^2 + \gamma \right) + (d-\kappa) \log \gamma, \nonumber
\end{eqnarray}
noting that this minimization was accomplished without encountering any suboptimal local minima.


\subsection{Optimization over parameters $\theta$ that are unique to $\bar{\ell}_x(\theta)$} \label{sec:Lbar_x_sup}

The optimal $\bb_x$ is just the convex maximum likelihood estimator given by the mean
\begin{equation} \label{eq:mean_bx_sup}
\bb_x = \int \left( \bI - \bW_x \bW_y \right)\bx \rho_{gt}^x(d \bx) - \bW_x \bb_y = \left( \bI - \bW_x \bW_y \right)\bc - \bW_x \bb_y ,
\end{equation}
where the second equality follows from Definition \ref{def:affine_surjective_model} in the main text.  Plugging this value into (\ref{eq:vae_cost_tot_affine3_sup}) and applying a standard trace identity, we arrive at
\begin{equation} \label{eq:vae_cost_tot_affine4_sup}
\bar{\ell}_x(\theta)   \equiv    \mbox{tr}\left[ \bS_{\epsilon_x} \left( \bV_x \bV_x^{\top}  + \gamma \bI\right)^{-1} \right] + \sum_{k=1}^\kappa \log \left( \| \bv_{x,k}\|_2^2 + \gamma \right) + (d-\kappa) \log \gamma,
\end{equation}
where
\begin{equation}
\bS_{\epsilon_x} \triangleq \mbox{Cov}_{\rho_{gt}^x}\left[ \bepsilon_x \right] = \left( \bI - \bW_x \bW_y \right) \mbox{Cov}_{\rho_{gt}^x}\left[\bx \right] \left( \bI - \bW_x \bW_y \right)^\top.
\end{equation}

The remaining parameters $\{\bW_x,\bW_y,\bV_x \}$ are all shared with the $\by \rightarrow \hat{\bx} \rightarrow \hat{\by}$ cycle loss $\ell_y(\theta)$, so ostensibly we must include the full loss $\bar{\ell}_x(\theta) + \ell_y(\theta)$ when investigating local minima with respect to these parameters. However, there is one subtle exception that warrants further attention here.  More specifically, the loss $\ell_y(\theta)$ depends on $\bV_x$ only via the outer product $\bV_x \bV_x^\top$.  Consequently, if $\bV_x = \bar{\bU} \bar{\bLambda} \bar{\bV}^\top$ denotes the singular value decomposition of $\bV_x$, then $\ell_y(\theta)$ is independent of $\bar{\bV}$ since $\bV_x \bV_x^\top = \bar{\bU} \bar{\bLambda} \bar{\bLambda}^\top \bar{\bU}^\top$, noting that $\bar{\bLambda} \bar{\bLambda}^\top$ is just a square matrix with squared singular values along the diagonal.  It then follows that we can optimize $\bar{\ell}_x(\theta)$ over $\bar{\bV}$ without influencing $\ell_y(\theta)$.

To this end we have the following:
\begin{lemma} \label{lem:cost_conversion_sup}
At any minimizer (local or global) of $\bar{\ell}_x(\theta)$ with respect to $\bar{\bV}$, it follows that $\bar{\bV} = \bP$ for some permutation matrix $\bP$ and the corresponding loss satisfies
\begin{equation}
\bar{\ell}_x(\theta) = \mathrm{tr}\left[ \bS_{\epsilon_x}  \bSigma_{\epsilon_x}^{-1} \right] + \log\left| \bSigma_{\epsilon_x}  \right|, ~~ \mbox{where} ~ \bSigma_{\epsilon_x} \triangleq \bV_x \bV_x^{\top}  + \gamma \bI.
\end{equation}
\end{lemma}
This result follows (with minor modification) from \citep{dai2017hidden}[Corollary 3].  A related result also appears in \citep{lucas2019don}.

\subsection{Optimization over parameters $\theta$ that are unique to $\ell_y(\theta)$} \label{sec:Ly_parameters_sup}

Since $\by$ has zero mean per Definition \ref{def:affine_surjective_model}, the optimal $\bb_y$ is the convex maximum likelihood estimator satisfying $\bb_y = - \bW_y \bb_x$ (this assumes that $\bW_y \bb_x$ has not been absorbed into $\by$ as mentioned in the main text for notational simplicity).  This leads to
\begin{equation}
\ell_y(\theta) \equiv \mbox{tr}\left[ \bS_{\epsilon_y} \bSigma^{-1}_{\epsilon_y}  \right] + \log\left| \bSigma_{\epsilon_y} \right|, ~~\mbox{where} ~ \bS_{\epsilon_y}  \triangleq \left(\bI - \bW_y \bW_x\right)\left(\bI - \bW_y \bW_x\right)^\top
\end{equation}
and $\bSigma_{\epsilon_y}$ is defined in the main text.

\subsection{Optimizing the combined loss $\bar{\ell}_{cycle}(\theta)$} \label{sec:Lbar_all_sup}

The above results imply that we may now consider jointly optimizing the combined loss
\begin{equation} \label{eq:reduced_cvae_cost_sup}
\bar{\ell}_{cycle}(\theta) \triangleq \bar{\ell}_x(\theta) + \ell_y(\theta)
\end{equation}
over $\{\bW_x,\bW_y, \bV_x \bV_x^\top\}$; all other terms have already been optimized out of the model without encountering any suboptimal local minima.  To proceed, consider the distribution $\rho^{\hat{y}}_{gt}$ of
\begin{equation}
\hat{\by} = \bW_y \bx + \bb_y =  \bW_y \bA \by + \bW_y \bB \bu + \bW_y \bc + \bb_y.
\end{equation}
To satisfy the constraint the stipulated constraint $\rho^{\hat{y}}_{gt} = \rho^{y}_{gt}$ subject to the conditions of Definition \ref{def:affine_surjective_model}, it must be that $\bW_y \bA = \bI$ and $\bB \in \mbox{null}[\bW_y]$ (it will also be the case that $\bb_y = - \bW_y \bc$ to ensure that $\hat{\by}$ has zero mean).  From this we may conclude that
\begin{eqnarray}
\bS_{\epsilon_x} & = & \left( \bI - \bW_x \bW_y \right) \mbox{Cov}_{\rho_{gt}^x}\left[\bx \right] \left( \bI - \bW_x \bW_y \right)^\top \nonumber \\
& = & \left( \bI - \bW_x \bW_y \right) \left[\bA \bA^\top + \bB \bB^{\top} \right] \left( \bI - \bW_x \bW_y \right)^\top \\
& = & \left(\bA - \bW_x\right)\left( \bA - \bW_x\right)^\top + \bB \bB^\top, \nonumber
\end{eqnarray}
where the middle equality follows because $\by$ and $\bu$ are uncorrelated with identity covariance.  Furthermore, let $\widetilde{\bD}\in \mathbb{R}^{r_y \times r_x}$ denote any matrix such that $\widetilde{\bD}\bA = \bI$ and $\bB \in \mbox{null}[\widetilde{\bD}]$.  It then follows that $\bW_y$ must equal some such $\widetilde{\bD}$ and optimization of (\ref{eq:reduced_cvae_cost_sup}) over $\bW_x$ will involve simply minimizing
\begin{equation} \label{eq:reduced_cvae_cost2_sup}
\bar{\ell}_{cycle}(\theta) \equiv \mbox{tr}\left[ \left(\bA - \bW_x\right)\left( \bA - \bW_x\right)^\top   \bSigma_{\epsilon_x}^{-1} \right] + \mbox{tr}\left[ \left(\bI - \widetilde{\bD}\bW_x\right)\left(\bI - \widetilde{\bD}\bW_x\right)^\top \bSigma^{-1}_{\epsilon_y} \right] + C
\end{equation}
over $\bW_x$, where $C$ denotes all terms that are independent of $\bW_x$.  This is a convex problem with unique minimum at $\bW_x = \bA$.  Note that this choice sets the respective $\bW_x$-dependent terms to zero, the minimum possible value.  Plugging $\bW_x = \bA$ into (\ref{eq:reduced_cvae_cost2_sup}) and expanding the terms in $C$, we then arrive at the updated loss
\begin{eqnarray} \label{eq:reduced_cvae_cost3_sup}
\bar{\ell}_{cycle}(\theta) & \equiv & \mbox{tr}\left[ \bB \bB^\top   \bSigma_{\epsilon_x}^{-1} \right] + \log\left|\bSigma_{\epsilon_x} \right| + \log\left| \bSigma_{\epsilon_y} \right| \\
& = & \mbox{tr}\left[ \bB \bB^\top   \left(\bV_x \bV_x^\top + \gamma \bI \right)^{-1} \right] + \log\left|\bV_x \bV_x^\top + \gamma \bI \right| + \log\left| \widetilde{\bD}\bV_x \bV_x^\top \widetilde{\bD}^\top + \gamma \bI \right|. \nonumber
\end{eqnarray}
Minimization of this expression over $\bV_x$ as $\gamma$ becomes arbitrarily small can be handled as follows.  If any $\bV_x$ and $\gamma$ are a local minima of (\ref{eq:reduced_cvae_cost3_sup}), then $\{\alpha = 1,\beta = 0\}$ must also be a local minimum of
\begin{eqnarray} \label{eq:ab_equation_sup}
& & \hspace*{-0.8cm} \bar{\ell}_{cycle}(\alpha,\beta) \triangleq \\
& &\mbox{tr}\left[ \bB \bB^\top \left( \alpha \bSigma_{\epsilon_x} + \beta \bB \bB^\top \right)^{-1} \right] + \log\left| \alpha \bSigma_{\epsilon_x} + \beta \bB \bB^\top \right| + \log\left| \alpha \bSigma_{\epsilon_y} + \beta \widetilde{\bD}\bB \bB^\top \widetilde{\bD}^\top \right| \nonumber \\
& &=~\mbox{tr}\left[ \bB \bB^\top \left( \alpha \bSigma_{\epsilon_x} + \beta \bB \bB^\top \right)^{-1} \right] + \log\left| \alpha \bSigma_{\epsilon_x} + \beta \bB \bB^\top \right| + \log\left| \alpha \bSigma_{\epsilon_y} \right|. \nonumber
\end{eqnarray}
If we exclude the second log-det term, then it has been shown in \citep{Wipf07d} that loss functions in the form of (\ref{eq:ab_equation_sup}) have a monotonically decreasing path to a unique minimum as $\beta \rightarrow 1$ and $\alpha \rightarrow 0$ .  However, given that the second log-det term is a monotonically decreasing function of $\alpha$, it follows that the entire loss from (\ref{eq:ab_equation_sup}) has a unique minimum as $\beta \rightarrow 1$ and $\alpha \rightarrow 0$.  Consequently, it must be that at any local minimum of (\ref{eq:reduced_cvae_cost3_sup}) $\bV_x \bV_x^\top = \bB\bB^\top$ in the limit as $\gamma \rightarrow 0$.  Moreover, the feasibility of this limiting equality is guaranteed by our assumption that $r_z \geq r_c - r_y$ (i.e., if $r_z < r_c - r_y$, then $\bV_x$ would not have sufficient dimensionality to allow $\bV_x \bV_x^\top = \bB\bB^\top$).

\subsection{Final Pieces}
We have already established that at any local minimizer $\{\theta^*,\phi^*\}$ it must be the case that $\bW_x^* = \bA$ and $\bW_y^* = \widetilde{\bD}$.  Moreover, we also can infer from (\ref{eq:mean_bx_sup}) and Section \ref{sec:Ly_parameters_sup} that at any local minimum we have
\begin{equation}
\bb_x^* = \left(\bI - \bW_x^* \bW_y^*\right) \bc - \bW^*_x \bb^*_y = \left(\bI - \bW_x^* \bW_y^*\right) \bc + \bW^*_x \bW_y^* \bb_x^* = \left(\bI - \bA \widetilde{\bD}\right) \bc + \bA \widetilde{\bD}\bb^*_x
\end{equation}
from which it follows that $\left(\bI - \bA \widetilde{\bD}\right)\bc = \left(\bI - \bA \widetilde{\bD}\right) \bb_x^*$.  This along is not sufficient to guarantee that $\bb_x^* = \bc$ is the unique solution; however, once we include the additional constraint $\rho_{gt}^y = \rho_\theta^{\hat{y}}$ per the Proposition \ref{prop:no_bad_local_min} statement, then $\bb_x^* = \bc$ is uniquely determined (otherwise it would imply that $\hat{\by}$ has a nonzero mean).  It then follows that $\bb_y^* = -\bW_y^* \bb_x^* = -\widetilde{\bD}\bc$.

And finally, regarding $\bV_x^*$, from Section \ref{sec:Lbar_all_sup} we have that $\bV_x^* \left( \bV_x^* \right)^\top = \bB \bB^\top$.  Although this does \emph{not} ensure that $\bV_x^* = \bB$, we can conclude that $\mbox{span}[\bar{\bU}] = \mbox{span}[\bB]$.  Furthermore, we know from Lemma \ref{lem:cost_conversion_sup} and the attendant singular value decomposition that $\bV_x^* = \bar{\bU} \bar{\bLambda} \bP^\top$ and $\left( \bV_x^* \right)^\top \bV_x^* = \bP^\top \bar{\bLambda}^\top \bar{\bLambda} \bP$.  Therefore, up to an arbitrary permutation, each column of $\bV_x^*$ satisfies
\begin{equation}
\| \bv^*_{x,k} \|_2^2 ~~ = ~~  \left\{ \begin{array}{ll} \bar{\lambda}_k^2, & \forall ~ k = 1,\ldots, \mbox{rank}[\bB] \\ 0, & \forall ~ k = \mbox{rank}[\bB]+1,\ldots, r_z\end{array} \right.
\end{equation}
where $\bar{\lambda}_k$ is an eigenvalue of $\bar{\bLambda}$.  Collectively then, these results imply that $\bV^*_x = \left[\widetilde{\bB}, {\bf 0} \right] \bP^\top$, where $\widetilde{\bB} \in \mathbb{R}^{r_x \times \mbox{rank}[\bB] }$ satisfies $\mbox{span}[\widetilde{\bB}] = \mbox{span}[\bU] = \mbox{span}[\bB]$.  


\section{Proof of Corollary \ref{cor:implicit_bijection}}

From (\ref{eq:mu_z_opt}) in the proof of Proposition \ref{prop:no_bad_local_min} and the derivations above, we have that at any optimal encoder solution $\phi^* = \{\bW^*_z,\bb_z^*\}$, both $\bW_z^*$ and $\bb_z^*$ are formed by left multiplication by $\left( \bV_x^* \right)^\top$.  Then based on Proposition \ref{prop:no_bad_local_min} and the stated structure of $\bV_x^*$, it follows that $\bW_z^* = \bP \left[ \begin{array}{c} \widetilde{\bW}_z^*  \\ {\bf 0} \end{array} \right]$ and $\bb_z^* = \bP \left[ \begin{array}{c} \widetilde{\bb}_z^*  \\ {\bf 0} \end{array} \right]$, where $\widetilde{\bW}_z^*$ has $\mbox{rank}[\bB ]$ rows and $\widetilde{\bb}_z^* \in \mathbb{R}^{\mbox{rank}[\bB ]}$.  Finally, there exists a bijection between $\bx$ and $\{\by,\widetilde{\bmu}_z\}$ given that
\begin{eqnarray}
& \by = \bW^*_y \bx + \bb_y^* ~\mbox{and}~ \widetilde{\bmu}_z = \widetilde{\bW}_z^*\bx + \widetilde{\bb}_z^*~~\left(\mbox{for} ~ \bx \rightarrow \{\by,\widetilde{\bmu}_z\} ~ \mbox{direction} \right) & \nonumber \\
& \bx = \bW_x^* \by + \bV_x^* \bP \left[ \begin{array}{c} \widetilde{\bmu}_z  \\ {\bf 0} \end{array} \right] + \bc~~\left(\mbox{for} ~  \{\by,\widetilde{\bmu}_z\} \rightarrow \bx ~ \mbox{direction} \right), &
\end{eqnarray}
completing the proof.

\end{document}